\documentclass[letterpaper, 10 pt, conference]{ieeeconf}
\IEEEoverridecommandlockouts

\usepackage{cite}

\usepackage{url}

\usepackage{amsmath,amssymb,amsfonts}
\usepackage{graphicx}
\graphicspath{ {./images/} }

\usepackage{algorithm}
\usepackage{algorithmic}
\usepackage{footnote}

\usepackage{tabularx, booktabs}
\usepackage{multirow}
\usepackage[footnotesize]{caption}

\usepackage[caption=false, font=footnotesize]{subfig}

\usepackage{etoolbox}

\usepackage{gensymb}
\usepackage{textcomp}
\usepackage{xcolor}

\def\BibTeX{{\rm B\kern-.05em{\sc i\kern-.025em b}\kern-.08em
    T\kern-.1667em\lower.7ex\hbox{E}\kern-.125emX}}
\begin{document}



\title{\LARGE \bf Autonomous Surface Selection For Manipulator-Based UV Disinfection In Hospitals Using Foundation Models \\
}

\author{Xueyan Oh$^{1}$, Jonathan Her$^{1}$, Zhixiang Ong$^{1}$, Brandon Koh$^{2}$, Yun Hann Tan$^{3}$, and U-Xuan Tan$^{1}$
\thanks{$^{1}$Authors are with the Engineering Product Development Pillar, Singapore University of Technology and Design, Singapore
        {\tt\footnotesize xueyan\_oh@sutd.edu.sg, uxuan\_tan@sutd.edu.sg}}
\thanks{$^{2}$Author is with Centre for Healthcare Assistive \& Robotics Technology, Singapore}
\thanks{$^{3}$Author is with National Centre for Infectious Diseases, Singapore}%
}

\makeatletter
\patchcmd{\@maketitle}
  {\addvspace{0.5\baselineskip}\egroup}
  {\addvspace{-2\baselineskip}\egroup}
  {}
  {}
\makeatother

\maketitle

\begin{abstract}
Ultraviolet (UV) germicidal radiation is an established non-contact method for surface disinfection in medical environments. Traditional approaches require substantial human intervention to define disinfection areas, complicating automation, while deep learning-based methods often need extensive fine-tuning and large datasets, which can be impractical for large-scale deployment. Additionally, these methods often do not address scene understanding for partial surface disinfection, which is crucial for avoiding unintended UV exposure. We propose a solution that leverages foundation models to simplify surface selection for manipulator-based UV disinfection, reducing human involvement and removing the need for model training. Additionally, we propose a VLM-assisted segmentation refinement to detect and exclude thin and small non-target objects, showing that this reduces mis-segmentation errors. Our approach achieves over 92\% success rate in correctly segmenting target and non-target surfaces, and real-world experiments with a manipulator and simulated UV light demonstrate its practical potential for real-world applications.
\end{abstract}



\section{Introduction}
The use of ultraviolet (UV) germicidal radiation as a non-contact approach for disinfection is well known and there is ample research in recent years that have proven their effectiveness to sterilise surfaces in medical environments \cite{uvdist, faruvcmm}, especially since the COVID-19 pandemic. There has been large efforts to automate the UV surface disinfection process using robots due to factors such as the risk that UV radiation poses to nearby humans, being a labor-intensive process, and the need for precise and consistent operation to deliver the appropriate dosage to surfaces \cite{uvdist}.

A common type of UV disinfection robot uses high-powered, uncovered UV lamps mounted vertically onto a mobile base that is either being teleoperated or navigates autonomously to desired locations. Examples include commercially available UV disinfection robots such as Tru-D SmartUVC \cite{trud} and Lightstrike by Xenex Disinfection Services \cite{xenex}, as well as those in research \cite{designdeployuvc, uvdistindoor, uvcpurge}. This approach has several disadvantages including the high risk uncovered UV lamps pose to nearby humans and that many high touch surfaces do not receive sufficient UV exposure due to shadowing \cite{uvdist} and the unfavourable orientation of UV rays on horizontal surfaces \cite{sharedautomm}. 

\begin{figure}[t]
\centerline{\includegraphics[width=\columnwidth]{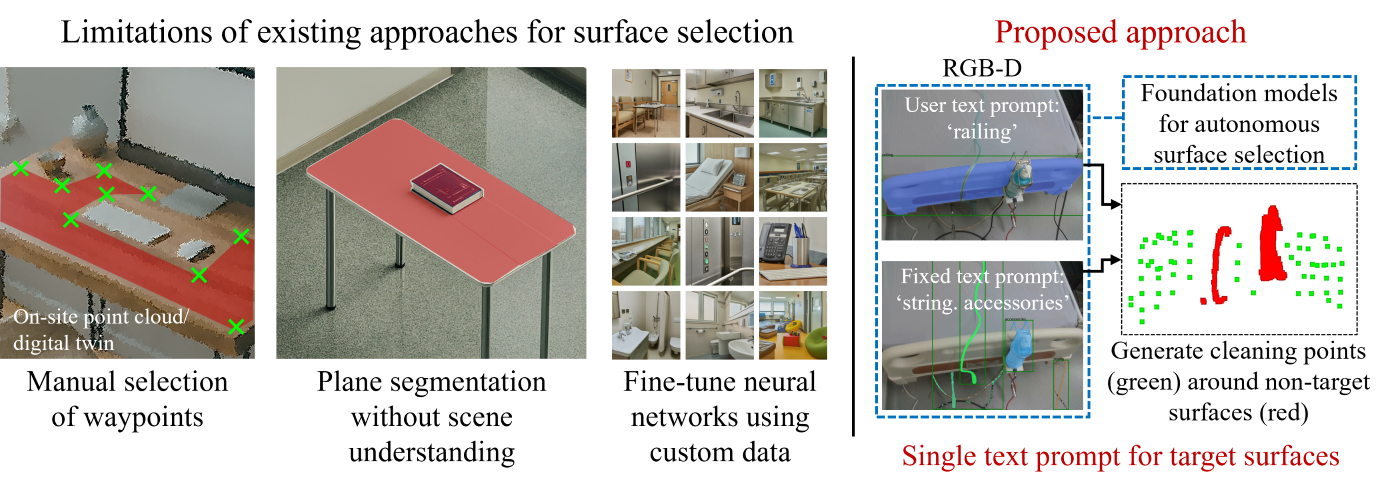}}
\vspace{-2mm}
\caption{Summary of the limitations in existing approaches and our proposed approach for autonomous surface selection and target point generation for manipulator-based UV disinfection in hospitals.}
\label{fig:problem}
\vspace{-6mm}
\end{figure}

Alternatively, other approaches \cite{ADAMMS, mapbaseduv, sharedautomm, faruvcmm, uvcrobotdev} address these limitations by using a UV light source as the manipulator's end effector, leveraging the additional degrees of freedom for precise positioning and orientation of the UV light source to optimise UV exposure. The ADAMMS-UV robot \cite{ADAMMS} is a mobile manipulator with a UV wand as its end effector that disinfects target surfaces selected by an operator through a GUI. Conte et al. \cite{mapbaseduv} uses a teleoperated mobile manipulator with UV lamps mounted to the sides of its base to disinfect floors and vertical surfaces, as well as to its end effector for horizontal surfaces. Sanchez and Smart \cite{sharedautomm} use a Fetch mobile manipulator with a UV flashlight for surface disinfection and offers a GUI that displays a voxel-based model of the environment. An operator can specify 3D points within this world to denote a horizontal area for disinfection and a path and motion plan is generated. Mehta et al. \cite{faruvcmm} uses a plane segmentation module on point cloud data to extract planar surfaces for UV disinfection by a manipulator but is limited to planar surfaces and without scene understanding. Ma et al. \cite{uvcrobotdev} propose a system where users manually select target areas through a GUI for mobile manipulator UV disinfection. A common limitation of the above approaches is the need for substantial human intervention to define surfaces for disinfection which is time-consuming and not practical for large-scale deployment. Furthermore, these works do not explore autonomous scene understanding for partial surface disinfection, which is necessary when objects such as medical equipment and personal belongings on or near to the target surface should be avoided to reduce the risk of unintended interaction. Objects with surface materials that reflect \cite{uvreflect} or degrade \cite{materialdegrade} under UV exposure also need to be shielded from UV light while surrounding surfaces are treated. In such cases, it is essential to ensure that the UV light only shines on the intended surfaces to minimise the risk of UV irradiation to surrounding areas, which is an important consideration when conducting UV disinfection in environments with human presence \cite{uvdist}.

Research has demonstrated the potential of deep learning models to detect and segment surfaces of common objects for cleaning applications \cite{segconareas, robothandles, recognsurface, deeplearningcleansurface}. However, these models often require substantial fine-tuning when applied to previously unseen objects, leading to increased development time and cost. On the other hand, foundation models (FMs) \cite{robotfm} are large, pre-trained models that can generalise well to new tasks with minimal or no fine-tuning due to the scale and diversity of data they are trained on, and the recent rise of FMs has created new opportunities in robotics. In this work, we leverage the generalisation capability of foundation models to address the challenges of surface selection in manipulator-based UV disinfection. Additionally, we propose a segmentation refinement approach to address mis-segmentation errors commonly encountered in image segmentation pipelines. Fig.~\ref{fig:problem} summarises our proposed approach and our main contributions are as follows:

\begin{itemize}
\item We propose a pipeline that leverages foundation models to autonomously extract cleaning points for manipulator-based UV surface disinfection. This eliminates the need to fine-tune deep learning models and simplifies the automation process by substantially reducing the need for human intervention in surface detection and selection.

\item We propose a VLM-assisted segmentation refinement that detects thin and small non-target objects to refine the non-target mask and reduce mis-segmentation errors. We evaluate its effectiveness on high-touch hospital surfaces and show that it improves exclusion of non-target surfaces compared to a baseline.

\item We demonstrate manipulator-based simulated UV disinfection using target cleaning points generated by our pipeline, showing its potential for practical applications.
\end{itemize}




\section{Related Work}

\subsection{Deep Learning-Based Object Detection and Segmentation For Cleaning Applications}\label{dldetseg}

Many recent works on autonomous cleaning robots leverage deep learning for detecting or segmenting relevant objects and surfaces. For example, Ramalingam et al. \cite{robothandles} propose a framework for automating cleaning of door handles that uses a Convolutional Neural Network (CNN) trained on custom data to specifically detect different types of door handles from images. The detected bounding box is aligned with depth data to obtain a region in 3D space to clean by spraying a liquid disinfectant and wiping using a cleaning brush. Qi et al. \cite{deeplearningcleansurface} propose a network that uses RGB-D data to generate a semantic segmentation of the input image, identifying regions that require cleaning. Their model is trained on a custom dataset, comprising pre-identified objects from their laboratory environment that need to be cleaned. Hu et al. \cite{segconareas} segment high-touch surfaces for cleaning using the concept of object affordance, where a network is trained to associate surfaces with specific human interactions using labels such as 'sit,' 'pull,' 'grasp,' and 'place' to segment surfaces based on their affordances. Rather than detect the type of object, Hu et al. \cite{recognsurface} propose a network that can classify surfaces by their material type, such as fabric, wood or metal, enabling adaptive disinfection methods and parameters. A common limitation of these approaches is their requirement for training or fine-tuning a network when applied to previously unseen objects or surfaces.

\subsection{Foundation Models for Object Detection and Segmentation in Robotics}\label{fmdetseg}

Foundation models, trained on large datasets, can be applied to a broad range of downstream tasks through context-based learning, fine-tuning, or even without additional training \cite{robotfm}. Recent Vision Language Models (VLMs) such as GLIP \cite{glip} and Grounding DINO \cite{groundingdino} enable zero-shot, open-vocabulary detection of objects from images based on text inputs. Similarly, Large Vision Models (LVMs) such as the Segment Anything Model (SAM) \cite{sam} and FastSAM \cite{fastsam} have achieved state-of-the-art performance in zero-shot image segmentation. Notably, Ghazouali et al. \cite{fusionvision} introduce FusionVision which integrates an object detector with SAM and RGB-D data for 3D reconstruction of detected objects, where their object detector can potentially be replaced with a VLM to detect unseen objects. However, methods combining object detection with SAM often face mis-segmentation challenges, particularly in cluttered environments where fine-grained or overlapping details lead to segmentation errors \cite{sam_micro, beyondhumanvision,samsurvey,uoissam}. Methods to address this include fine-tuning models \cite{sam_micro} and applying refinement steps to reduce background mis-segmentation \cite{uoissam}. Our approach reduces mis-segmentation of thin and small objects placed over target surfaces by specifically detecting these non-target objects and excluding them from the target surfaces. While foundation models have been incorporated into diverse robotics applications \cite{fmrobotics,anomaly,tunnel}, their use in autonomous cleaning tasks remains underexplored.

\section{Proposed Pipeline}

\subsection{Overview of Pipeline}

\begin{figure*}[t]
\centering
\centerline{\includegraphics[width=1.0\textwidth]{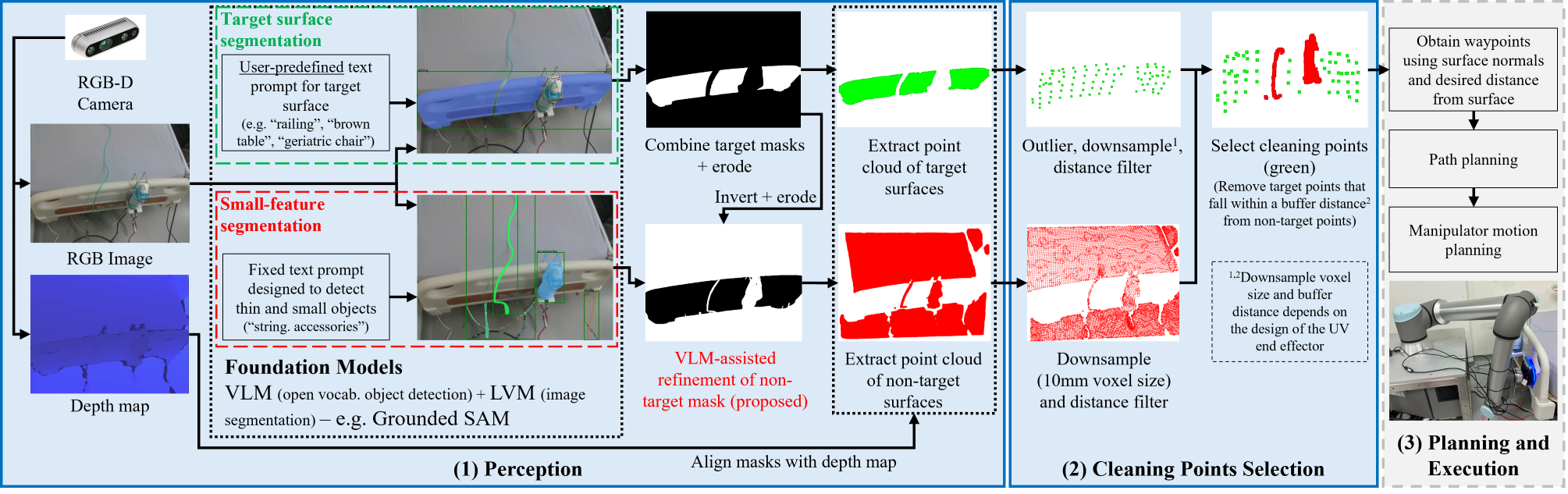}}
\vspace{-1mm}
\caption{Overview of our proposed pipeline that leverages foundation models for autonomous perception and selection of cleaning points for manipulator-based UV disinfection of high-touch surfaces in hospitals. The pipeline consists of (1) A Perception module that passes an RGB image twice into Grounded SAM---first with a user-defined prompt to segment target surfaces, and second with a fixed prompt to detect thin and small objects for our proposed VLM-assisted refinement of the non-target mask, reducing mis-segmentation errors in the target mask. The resulting masks are aligned with the depth map to extract point clouds; (2) A Cleaning Points Selection module processes these point clouds using a series of filters to select the final surface points for UV disinfection; 3) The Planning and Execution module can use these selected points to generate waypoints and the path and motion a manipulator. Modules 1 and 2 in blue represent our main contributions while module 3 is included for completeness.}
\label{fig:pipeline}
\vspace{-6mm}
\end{figure*}

We propose a pipeline capable of autonomously generating waypoints for manipulator-based UV disinfection of high-touch surfaces in hospital environments. This pipeline leverages the capabilities of state-of-the-art foundation models to remove the requirement for human input such as manually selecting surfaces to disinfect or training a network to detect appropriate high-touch surfaces. Our approach can also exclude non-target surfaces that should be avoided, preventing unnecessary UV exposure and minimizing risks to nearby humans. Designed mainly as Perception and Cleaning Point Selection modules, this pipeline can be easily integrated with downstream processes such as path planning and motion planning in autonomous manipulator-based UV disinfection systems. We assume that an RGB-D camera, which can be mounted on a mast or the wrist joint of a manipuator, is overlooking the manipulator's workspace while the object to be cleaned is in the camera view and within reach of the manipulator. This can be achieved by either bringing the target object to the front of the disinfection system, such as a cleaning kiosk, or using a mobile manipulator that can navigate to a waypoint near to and facing the target object.

Our pipeline consists of the following modules: 1) Perception: An RGB image, obtained from an RGB-D camera, and a predefined text prompt consisting of object labels are input into a VLM to predict bounding boxes of relevant objects. These bounding boxes, along with the original RGB image, are then passed to a LVM for image segmentation, generating a target mask and a non-target mask. We propose to explicitly detect and segment thin and small non-target objects to refine the non-target mask. The resulting segmentation masks are aligned with the depth map to extract point clouds of the target and non-target surfaces. 2) Cleaning Points Selection: The extracted point clouds are processed to obtain the final selection of points for UV disinfection. 3) Planning and Execution: The selected cleaning points are used for path and motion planning of a manipulator with a UV disinfection end effector. The Perception and Cleaning Points Selection modules are our main contributions while the Planning and Execution module is included for completeness. Fig.~\ref{fig:pipeline} presents an overview of our proposed pipeline, and we explain each module in detail in the following sections.

\subsection{Perception}\label{perception}
In the Perception module, we leverage two visual foundation models (FMs) to generate segmentation masks relevant to our cleaning task. The first FM is a VLM capable of using a text prompt to detect arbituary objects within an image and we use Grounding Dino \cite{groundingdino}, which combines transformer-based object detection with grounded pre-training, for its state-of-the-art performance in open vocabulary object detection on benchmark datasets. The VLM uses a text prompt and an RGB image as inputs to predict one or more bounding boxes locating the target object(s). The second FM is a LVM for image segmentation and we use SAM \cite{sam} for its outstanding zero-shot performance and capability of accepting a bounding box as a prompt for instance segmentation. Each bounding box obtained from the output of Ground DINO is used as a prompt into SAM, along with the original RGB image, to obtain segmentation masks. For implementation, we use Grounded SAM \cite{groundedsam} which combines Grounding DINO and SAM in this manner.

\textit{1) Obtaining Target Mask:} We pass the RGB image twice into Grounded SAM. The first pass aims to output segmentation masks of the target surfaces by using a predefined text prompt and a confidence threshold of 0.35. We assume that the target object with high-touch surfaces requiring UV disinfection is known prior to commencing the cleaning operation and the user is required to predefine a simple text prompt that describes the object. Users can refer to an article by Skalski \cite{groundingdinoblog} for guidelines on choosing prompts and it is recommended to test the text prompt on several images of their target object to ensure that the relevant surfaces can be detected prior to deployment. If there are multiple masks, we combine them to form the final target mask. To filter segmentation and sensor noise, we apply erosion with OpenCV's erode function using a 10x10 kernel of ones. However, mis-segmentation errors are often still present and we propose to reduce these errors by using a non-target mask.

\begin{figure}[t]
\centerline{\includegraphics[width=\columnwidth]{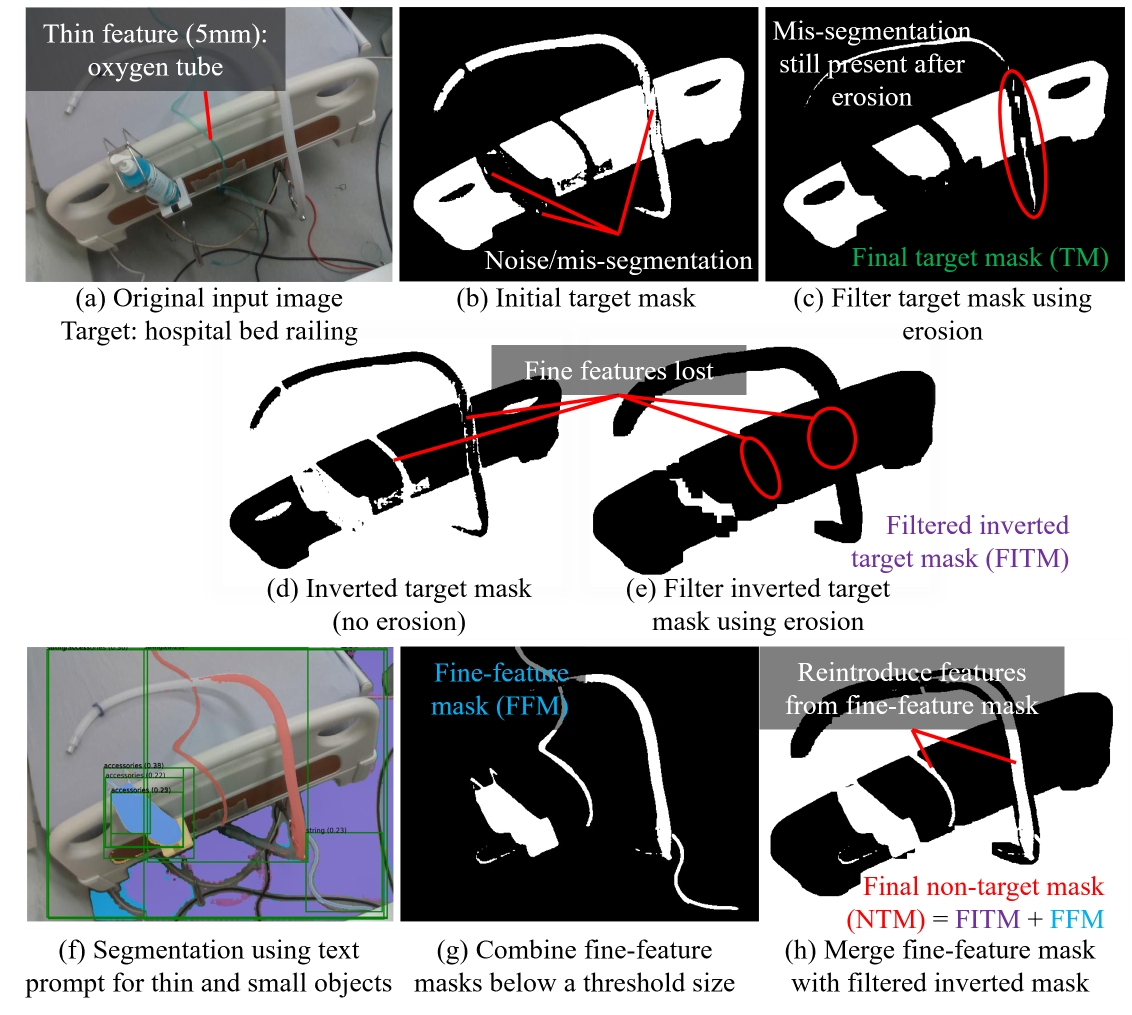}}
\vspace{-2mm}
\caption{Applying erosion using a 10x10 kernel of ones to the initial target mask (b) effectively reduces noise but mis-segmentation errors are still present (c). A 20x20 kernel is required to sufficiently filter noise from the inverted target mask (d) which represents non-target surfaces, but this often leads to the loss of fine features (e) such as an oxygen tube. We address this by obtaining a fine-feature mask using a separate prompt designed to detect thin and small objects for segmentation (f). The output masks are then combined to obtain a fine-feature mask (g), which is then merged with the filtered inverted target mask to obtain the final non-target mask (h).}
\label{fig:erosion}
\vspace{-5mm}
\end{figure}

\textit{2) Obtaining Non-Target Mask With VLM-Assisted Refinement:} A separate mask that represents surfaces less the target surface by inverting the unfiltered target mask to obtain an inverted mask representing all non-target surfaces. We find that applying erosion with a 20x20 kernel to this inverted mask is essential to remove noise. However, this often leads to loss of fine features and we address this by passing the initial RGB image into Grounded SAM for a second time to detect and segment thin and small objects in order to isolate their masks. This step does not require any human input as we use a fixed prompt, 'string. accessories' to detect thin and small objects, and we use a confidence threshold of 0.2. In particular, we empirically find that 'string' and 'accessories' are more robust prompts to detect thin and small objects respectively, compared to more generic prompts such as 'thin objects' or 'small objects'. We combine all output masks with a mask size lower than 20,000 ones (experimentally determined threshold) to obtain a fine-feature mask and do not apply erosion. This fine-feature mask is then merged with the filtered inverted mask to reintroduce the fine features, forming the final non-target mask. Fig.~\ref{fig:erosion} uses a sample image to show the masks obtained and the effects of erosion. The final target and non-target masks are aligned with the depth map to extract a target and a non-target point cloud.

\subsection{Cleaning Points Selection}\label{cleaningpoints}
The two extracted point clouds are first transformed to the base frame of the manipulator before passing through a distance filter to remove points that are out of the manipulator's reach. We then apply statistical outlier removal using the parameters suggested in \cite{fusionvision} to the filtered target point cloud. These outliers are typically due to sensor noise or differences between the predicted segmentation masks from the Perception module and the ideal mask of each surface, leading to extraction of 3D points that are incorrectly labelled as the target surface. This is not applied to the non-target point cloud to avoid unintentionally removing fine features.

These two filtered point clouds are downsampled using a voxel grid filter while keeping the nearest detected point to each voxel's centroid. This ensures that the downsampled point cloud is a subset of the original observed point cloud which is essential for precision in our cleaning application. We use a voxel size of $v_{nt}=10$ (corresponding to about 10mm in the real world) for downsampling the points of non-target surfaces to reduce computational requirements for subsequent processes.

On the other hand, the voxel size, $v_t$, for downsampling the points of target surfaces is chosen based on the UV end effector's light distribution and intended distance from the target surface as this determines the spacing between target points for disinfection. While choosing $v_t$ to be much smaller than the area of UV exposure ensures overlapping areas exposed to UV, this can lead to excessive UV dosage to many areas and substantially increases the total time to disinfect the entire the surface. In view of minimising exposure to unintended surfaces, it is assumed that the end effector is compact and emits UV light that is relatively directional and has a high order of symmetry, such as the circular UV flashlight used in \cite{sharedautomm} and the square UV module in \cite{faruvcmm}. This enables choosing a $v_t$ value that is between half and the full the length or diameter of the end effector. While optimising the time taken for task completion is not our main contribution, using this approach to choose $v_t$ can offer some balance in efficiency and UV coverage but users are encouraged to experiment further to refine this.

Finally, we remove points from the downsampled target point cloud that fall within a distance (also using $v_t$) from any point in the non-target point cloud. This creates a buffer zone where target points that are too near to non-target points are omitted so that they will not be considered in downstream processes such as waypoint generation, reducing the chance for unintended interactions. The remaining points form the final set of selected cleaning points for UV disinfection. The pseudo code summarising the described processes for cleaning points selection is given in Algorithm~\ref{alg:cleaningpoints}.

\begin{algorithm}[t]
\caption{Cleaning Points Selection Algorithm}
\label{alg:cleaningpoints}
\begin{algorithmic}[1]
\STATE \textbf{Input:} Target point cloud $P_t$, Non-target point cloud $P_{nt}$, Voxel sizes $v_{nt}$, $v_t$
\STATE \textbf{Output:} Final selected cleaning points $P_{clean}$

\COMMENT{Transform and filter points based on reach}
\STATE $P_t, P_{nt} \leftarrow \text{TransformAndFilter}(P_t, P_{nt},$ \\
\hspace{12mm} $\text{Base Frame}, \text{Max Reach})$

\COMMENT{Outlier removal and downsampling of point clouds}
\STATE $P_t \leftarrow \text{StatisticalOutlierRemoval}(P_t)$ \cite{fusionvision}
\STATE $P_t \leftarrow \text{VoxelGridFilter}(P_t, v_t)$
\STATE $P_{nt} \leftarrow \text{VoxelGridFilter}(P_{nt}, v_{nt})$

\COMMENT{Remove target points near non-target points}
\STATE $P_t \leftarrow P_t - \{p \in P_t \mid \exists p_{nt} \in P_{nt}, \text{Distance}(p, p_{nt}) < v_t\}$

\STATE \textbf{return} $P_t$ as $P_{clean}$
\end{algorithmic}
\end{algorithm}

\subsection{Planning and Execution}\label{planning}
While the previous modules detail the main contributions of this paper, this section outlines the Planning and Execution component of our pipeline for completeness. UV disinfection is a non-contact task, requiring the UV source to be positioned at a predefined standoff distance (typically 1–2 inches) from the target surface and orienting it toward the selected points. Although exposure time and dosage are critical for disinfection, they depend on UV intensity and operational requirements and are beyond this paper’s scope. For positioning, we generate waypoints by applying an offset along the surface normals of the selected points, estimated using methods such as Open3D’s normal estimator.

Path planning optimises the order of waypoints for efficient disinfection coverage. This can be achieved using various common approaches, such as using a simple ziczag pattern or solving a Travelling Salesman Problem (TSP) to minimise overall travel distance, as demonstrated in \cite{faruvcmm, coverage}.

Motion planning generates valid joint trajectories to move the manipulator and its end effector to desired poses. This can be achieved using the ROS2 MoveIt framework with its integrated Open Motion Planning Library (OMPL), as discussed in \cite{faruvcmm, devandeva}. This approach also incorporates collision avoidance given environmental point cloud data for efficient and safe movement around obstacles.

\section{Experimental Evaluation}
\begin{figure}[t]
\centerline{\includegraphics[width=\columnwidth]{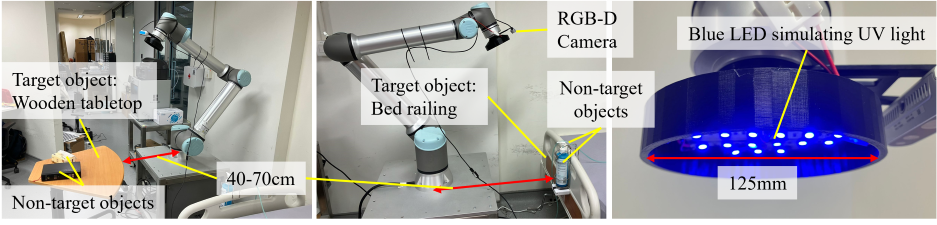}}
\vspace{-2mm}
\caption{Experimental setup featuring a UR10e manipulator with an RGB-D camera and a custom UV light-emitting end effector. In each experiment, the manipulator is positioned 40–70 cm from the nearest point of the target object, with the end effector initially oriented toward the target.}
\label{fig:setup}
\vspace{-3mm}
\end{figure}

Our experimental setup consists of a UR10e manipulator with a wrist-mounted Intel RealSense D435i RGB-D camera and a custom UV end effector. The end effector simulates UV light emission using blue LEDs mounted within a 125mm diameter circular area. We found that a voxel size and buffer distance of $v_t=70$, for the processes described in section \ref{cleaningpoints}, worked reasonably well across our experiments. The pipeline was excecuted on an NVIDA RTX 2080Ti GPU and tested on five target objects commonly found in hospitals, with relevant non-target objects (Table~\ref{tab:objects}). In all experiments, the manipulator's base was positioned between 40 to 70cm from each target object, simulating realistic distances for base placement and goal tolerance if using a mobile manipulator. Fig.~\ref{fig:setup} shows our experimental setup with a wooden tabletop and a bed railing as examples of target surfaces.
    

\captionsetup[table]{position=bottom}

\newcolumntype{C}{>{\centering\arraybackslash}X}
\newcolumntype{R}{>{\raggedright\arraybackslash}X}

\begin{table}[t]
\vspace{+0mm}
\begin{center}
\begin{tabularx}{1.0\columnwidth}{|m{2.4cm} |m{5.4cm}|}
\hline
\centering Target object/surface & \centering\arraybackslash Relevant non-target objects (to avoid) \\
\hline
\centering \rule{0pt}{3.2ex} White tabletop & \multirow{2}{=}{\centering\arraybackslash Ear thermometer, oximeter, eyeglasses, mobile phone, water bottle, thin orange book, thick black book, tissue pack} \\
\cline{1-1} 
\centering \rule{0pt}{3.2ex} Wooden tabletop & \\
\hline
\centering Bed railing/tailboard & \centering\arraybackslash Disinfectant with holder, oxygen tubes (thick, clear, green), drainage tube (orange)\\
\hline
\centering Black high chair & \multirow{2}{=}{\centering\arraybackslash Jacket, eyeglasses, mobile phone, water bottle, thin orange book, thick black book} \\
\cline{1-1} 
\centering Geriatric chair & \\
\hline

\end{tabularx}
\end{center}
\vspace{-3mm}
\caption{Target objects with high touch surfaces for UV disinfection and non-target objects in our experiments, which are common in hospitals.}
\vspace{-3mm}
\label{tab:objects}
\end{table}

\captionsetup[table]{position=bottom}

\newcolumntype{C}{>{\centering\arraybackslash}X}
\newcolumntype{R}{>{\raggedright\arraybackslash}X}

\begin{table}[t]
\vspace{+0mm}
\begin{center}
\begin{tabularx}{1.0\columnwidth}{m{2.4cm} m{1.8cm} m{0.7cm} m{0.8cm} m{0.8cm}}
\hline
\centering Target object/surface & \centering Target text prompt & \centering T(\%) & NT(\%) woNTM\centering & \centering\arraybackslash NT(\%) wNTM\\
\hline
\centering White tabletop & \centering ‘white table’ & \centering 100.0 & \centering 100.0 & \centering\arraybackslash \textbf{100.0} \\
\centering Wooden tabletop & \centering ‘brown table’ & \centering 100.0 & \centering 100.0 & \centering\arraybackslash \textbf{100.0} \\
\centering Bed railing/tailboard & \centering ‘railing’ & \centering 92.5 & \centering 89.2 & \centering\arraybackslash \textbf{93.3}\\
\centering Black highchair & \centering ‘black chair’ & \centering 96.3 & \centering 80.0 & \centering\arraybackslash \textbf{89.2} \\
\centering Geriatric chair & \centering ‘geriatric chair’ & \centering 73.8 & \centering 87.5 & \centering\arraybackslash \textbf{95.8} \\
\hline
\centering Overall & \centering - & \centering 92.5 & \centering 91.5 & \centering\arraybackslash \textbf{95.7} \\
\hline

\end{tabularx}
\end{center}
\vspace{-3mm}
\caption{Robustness of our perception module in successful segmentation of target surfaces (T) and exclusion of non-target surfaces (NT), comparing with (w) and without (wo) our proposed non-target mask (NTM). T scores remain unchanged regardless of whether NTM was used. A total of 400 unique images were captured, with 600 non-target object appearances.}
\vspace{-5mm}
\label{tab:segmentation}
\end{table}

\begin{figure*}[t]
\centerline{\includegraphics[width=1.0\textwidth]{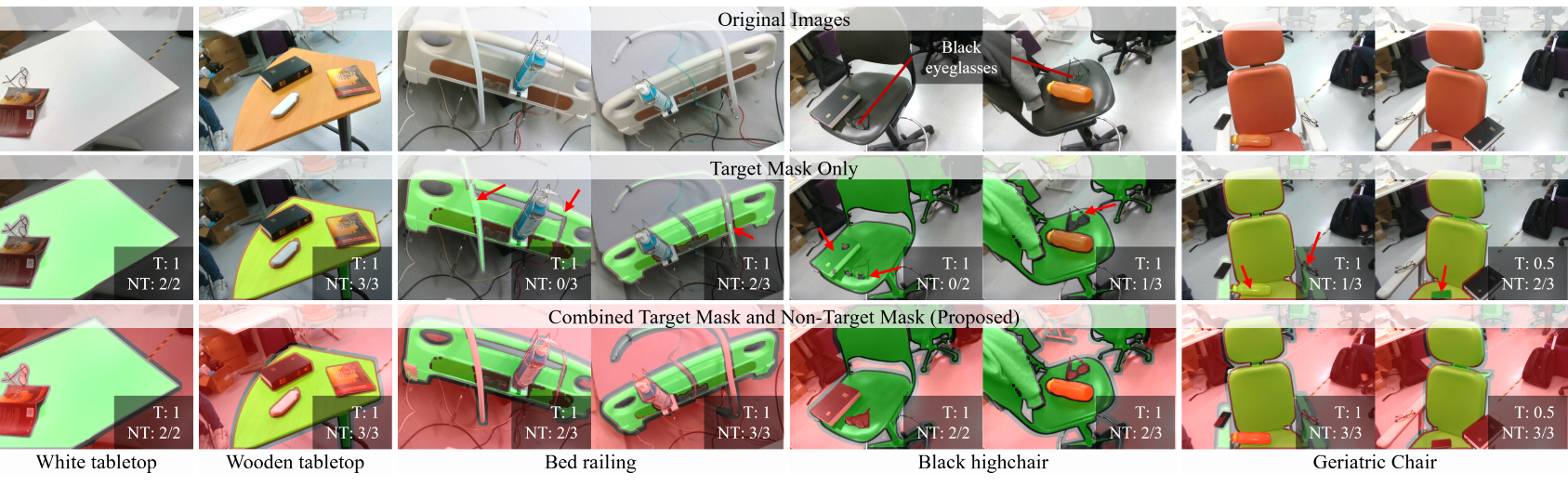}}
\vspace{-3mm}
\caption{Sample images, segmentation results in green (target mask) and red (non-target mask), and scores given for successful segmentation of target surfaces (T) and exclusion of non-target objects (NT, where 1/3 refers to 1 out of 3 objects successfully excluded). Combining our proposed non-target mask with the target mask effectively reduces mis-segmentation errors in the target mask (red arrows). Our pipeline is generally robust to various combinations of target surfaces and randomly placed non-target objects, even under high colour similarity. Possible failure modes include segmenting the geriatric chair's armrest as a target surface (geriatric chair, right) and failing to exclude the base of the disinfectant holder (bed railing, left).}
\label{fig:seg_results}
\vspace{-5mm}
\end{figure*}

\begin{figure*}[t]
\centering
\centerline{\includegraphics[width=1.0\textwidth]{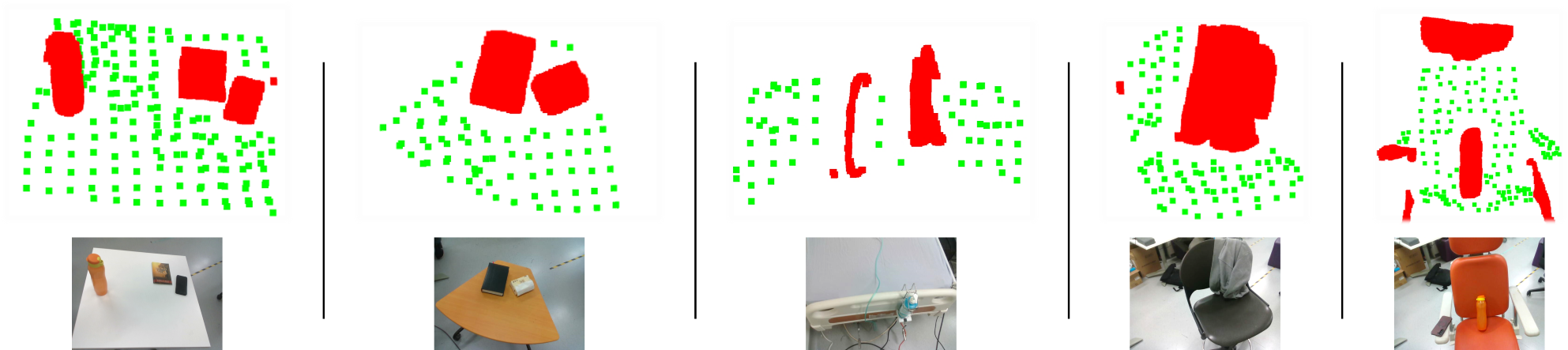}}
\vspace{-1mm}
\caption{Examples of successful selection of cleaning points for each target object. Green points represent target surfaces spaced approximately $v_t$mm (specified by the user) apart, indicating where the UV light should be directed. Red regions denote non-target surfaces that the UV end effector must avoid. The pipeline also generates a $v_t$mm buffer zone around these non-target surfaces, ensuring no cleaning points are placed nearby. These cleaning points are ready for use in downstream processes such as waypoint generation and enable precise and efficient UV targeting for disinfection by a manipulator, minimising the risk of unintended interactions and UV exposure to non-target surfaces.}
\label{fig:cleaningpoints}
\vspace{-6mm}
\end{figure*}

\subsection{Robustness of Target Surface Segmentation and Reducing Mis-segmentation Errors}
To assess the practicality of our approach for real-world manipulator-based UV disinfection, we evaluate its robustness in segmenting target surfaces and correctly excluding non-target surfaces in image space. Using the RGB-D camera, we captured 80 images of each of the five target objects from 20 different camera poses. For each pose, four images are taken with an increasing number of relevant non-target objects (0 to 3) placed on the target surface, as detailed in Table~\ref{tab:objects}. Since no ground truth data is available, we introduce a scoring metric where a target score (T) of 1 is assigned if at least half of the visible, unobstructed target surface is correctly segmented. A non-target score (NT) of 1 is assigned for each correctly excluded non-target object, provided no visible mis-segmentation occurs over that object. Unlike T, NT is scored more stringently, as downstream tasks may require complete avoidance of non-target objects to ensure precise disinfection. For the geriatric chair, where both the armrests and seat are critical high-touch surfaces, we treat them as separate target regions, assigning 0.5 points per correctly segmented surface. To ensure fairness, an independent researcher was briefed on the scoring criteria and assisted in evaluation.

We present our results in Table~\ref{tab:segmentation} along with the target text prompt used for each scene. Our pipeline achieves a success rate of over 92\% in detecting and segmenting various combinations of target and non-target surfaces, demonstrating its robustness across diverse high-touch surfaces and non-target objects commonly found in hospitals. The perfect scores for segmenting tabletop surfaces suggests that the FMs handle flat, homogeneous surfaces more easily compared to complex objects like bed railings and chairs. We show that combining our proposed non-target mask with the target mask improves exclusion of non-target surfaces by 4-11\% in challenging scenes (bed railings and chairs) compared to using only the target mask. Fig.~\ref{fig:seg_results} presents sample segmentation results and their scores, showing the pipeline’s ability to handle challenging scenarios. We show examples where our VLM-assisted segmentation refinement successfully detects and excludes non-target objects that were mis-segmented in the target mask, improving non-target exclusion.

We identify potential failure modes primarily due to color similarities or contrasts between target and non-target surfaces, and limitations in vision-based foundation models. For instance, the geriatric chair shows the lowest success rate in segmenting target surfaces, as armrests are difficult to detect due to their color contrast with the chair and similarity to the floor. Similarly, the grey disinfection holder's base, and thin or transparent objects such as oxygen tubes or spectacles frames, present challenges in excluding non-target surfaces. However, these failure modes may only reduce efficiency but not completely hinder disinfection tasks. Future work could address these challenges by exploring tailored prompts, incorporating point cloud-based foundation models, or combining segmentations from multiple viewpoints.

\subsection{Obtaining Final Selected Cleaning Points and Real World Demonstration Using a Manipulator}
\begin{figure}[t]
\centerline{\includegraphics[width=\columnwidth]{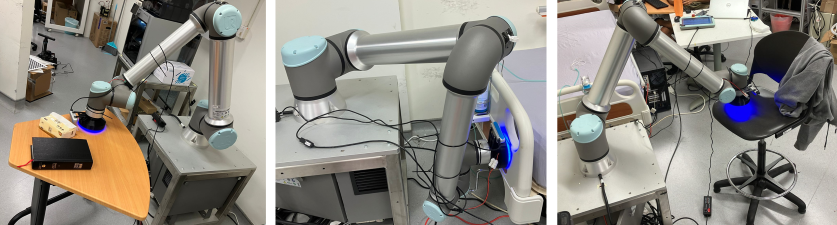}}
\vspace{-1mm}
\caption{Demonstrations of manipulator-based simulated UV disinfection of target surfaces while avoiding non-target objects, using waypoints generated from the cleaning points selected by our pipeline.}
\label{fig:demo}
\vspace{-5mm}
\end{figure}

We present successful selection of cleaning points (green) in Fig.~\ref{fig:cleaningpoints}, along with the creation of buffer zones around detected non-target surfaces (red) where no cleaning points are placed. These cleaning points are prepared for use in downstream processes, as described in section \ref{planning}, enabling precise and efficient UV targeting on high-touch surfaces while maintaining the specified distance from nearby non-target surfaces. This approach minimises the risk of unintended interactions and UV exposure to non-target surfaces, advancing manipulator-based UV disinfection in hospital environments with human presence. We also demonstrate manipulator-based simulated UV disinfection of target surfaces while avoiding non-target objects, by moving its end effector to waypoints generated from the cleaning points selected by our pipeline, with images shown in Fig.~\ref{fig:demo}.

\section{Conclusion}
In this work, we present a pipeline that utilises foundation models to autonomously select target surfaces and exclude non-target surfaces for manipulator-based UV disinfection. 3D points are extracted from the selected surfaces, with buffer zones created around non-target areas, preparing them for downstream processes such as waypoint generation and manipulator control. We also introduce a VLM-assisted segmentation refinement approach that uses a non-target mask to reduce mis-segmentation errors, improving exclusion of non-target surfaces. Our approach requires minimal operator input, with a simple text prompt to specify target surfaces before the cleaning operation, and eliminates the need for fine-tuning deep learning models or human intervention during disinfection. We demonstrate the robustness and practical potential of this method through real-world experiments, while also addressing its limitations and providing suggestions for future performance improvements.

\section*{Acknowledgment}

This research is supported by the National Robotics R\&D Programme Office, Singapore and Agency for Science Technology and Research (A*STAR), Singapore (M23NBK0024). The authors wish to thank the Centre for Healthcare Assistive \& Robotics Technology, Singapore, and the National Centre for Infectious Diseases, Singapore, for their support in conception and data acquisition for this study.

\bibliographystyle{./bibliography/IEEEtran}
\bibliography{./bibliography/main}

\end{document}